%% file: main.tex
\documentclass[11pt]{article}

\usepackage{microtype}
\usepackage{graphicx}
\usepackage{subfigure}
\usepackage{booktabs} %

\usepackage[margin=2.9cm]{geometry}
\PassOptionsToPackage{linktocpage}{hyperref}
\usepackage[hyperindex,breaklinks,colorlinks,linkcolor=blue,citecolor=blue,bookmarks=false,pagebackref]{hyperref}

\title{Kernel Methods and Multi-layer Perceptrons Learn Linear Models in High Dimensions}
\author{Mojtaba Sahraee-Ardakan$^{\dagger,\#}$, Melikasadat Emami$^\dagger$, Parthe Pandit$^{\ddag}$,\\[5pt] Sundeep Rangan$^\diamondsuit$, Alyson K. Fletcher$^{\dagger,\#}$\\[15pt]
\small$^\dagger$Department of Electrical and Computer Engineering, UCLA,\\
\small$^\#$Department of Statistics, UCLA, \\
\small$^\ddag$Hal{\i}c{\i}o\u{g}lu Data Science Institute, UCSD\\
\small$^\diamondsuit$Department of Electrical and Computer Engineering, NYU.}
\date{} 

\usepackage[round]{natbib}

\usepackage[utf8]{inputenc} %
\usepackage[T1]{fontenc}    %
\usepackage{hyperref}       %
\usepackage{url}            %
\usepackage{booktabs}       %
\usepackage{nicefrac}       %
\usepackage{microtype}      %

\usepackage{xcolor}
\usepackage{enumitem}
\usepackage{bbm}
\usepackage{amssymb,amsmath,amsfonts, amsthm, mathtools}
\usepackage{bm}

\usepackage{tikz}
\usetikzlibrary{arrows}

\input{commands.tex}

\input{commands_this_doc}

\usepackage{fancyhdr}
\usepackage{lastpage}

\renewcommand{\headrulewidth}{0pt}

\mathtoolsset{showonlyrefs=true}
\begin{document}
\maketitle

\input{sections/abstract}

\input{sections/intro_v3}

\input{sections/main_resutls}

\input{sections/proof_sketch}

\input{sections/experiments}

\input{sections/discussion}

\newpage
\bibliography{ref}
\bibliographystyle{apalike}

\input{sections/appendix}

\end{document}

%% file: commands.tex
\newcommand{\trace}[1]{{\rm tr}\left(#1\right)}

\newcommand\Ec{\mc E}

\newcommand{\Real}{\mathbb{R}}

\newcommand{\Exp}{\mathbb{E}}

\newcommand{\inv}{^{-1}}

\newcommand{\krr}{\mathsf{krr}}
\newcommand{\lin}{\mathsf{lin}}
\newcommand{\opt}{\mathsf{opt}}
\newcommand{\inner}[1]{\langle #1 \rangle}
\newcommand{\innerH}[1]{\langle #1 \rangle_{\mc H}}
\newcommand{\innerL}[1]{\langle #1 \rangle_{L^2}}

\newcommand{\tnorm}[1]{\|#1\|_2}

\DeclareMathOperator*{\argmin}{arg\,min }

\newcommand{\red}[1]{\textcolor{red}{#1}}

\newtheorem{theorem}{Theorem}[section]
\newtheorem{lemma}[theorem]{Lemma}
\newtheorem{proposition}[theorem]{Proposition}
\newtheorem{corollary}[theorem]{Corollary}

\theoremstyle{definition}
\newtheorem{remark}{Remark}

\newcommand{\mc}[1]{\mathcal{#1}}

\newcommand{\wh}[1]{\widehat{#1}}

\newcommand{\Mbf}{\mathbf{M}}
\newcommand{\Kbf}{\mathbf{K}}
\newcommand{\xbf}{\mathbf{x}}

\newcommand{\mubf}{\boldsymbol{\mu}}

\newcommand{\eqp}{\stackrel{\mathsf{p}}{=}}

\def\beq{\begin{equation}}
\def\eeq{\end{equation}}
\def\beqstar{\begin{equation*}}
\def\eeqstar{\end{equation*}}
\newcommand{\defeq}{\vcentcolon=}
\newcommand{\fhat}{\widehat{f}}

\newcommand{\mcF}{\mathcal{F}}

\newcommand{\ones}{\mathbf{1}}
\newcommand{\mcN}{\mathcal{N}}

\def\bm{\begin{matrix}}
\def\bbm{\begin{bmatrix}}
\def\ebm{\end{bmatrix}}

\newcommand{\Ibf}{\mathbf{I}}

\newcommand{\tran}{^{\text{\sf T}}}

\definecolor{ppink}{rgb}{0.858, 0.188, 0.478}

\providecommand{\ht}{\widetilde{h}}
\providecommand{\red}[1]{\textcolor{red}{#1}}
\RequirePackage{bm}

%% file: commands_this_doc.tex
\newcommand{\yhat}{\widehat{y}}

\def \tr{{\rm tr}}
\def \ts{{\rm ts}}

\newcommand{\what}{\widehat{w}}
\newcommand{\bhat}{\widehat{b}}

\newcommand{\mcH}{\mathcal{H}}

\newcommand{\fkhat}{\widehat{f}_{\rm ker}}

%% file: sections/abstract.tex
\begin{abstract}
Empirical observation of high dimensional phenomena, such as the double descent behaviour, has attracted a lot of interest in understanding classical techniques such as kernel methods, and their implications to explain generalization properties of neural networks. Many recent works analyze such models in a certain high-dimensional regime where the covariates are independent and the number of samples and the number of covariates grow at a fixed ratio (i.e.\ proportional asymptotics).
In this work we show that for a large class of kernels, including the neural tangent kernel of fully connected networks, kernel methods can only perform as well as linear models in this regime. 
More surprisingly, when the data is generated by a kernel model where the relationship between input and the response could be very nonlinear, we show that linear models are in fact optimal, i.e.\ linear models achieve the minimum risk among all models, linear or nonlinear.
These results suggest that more complex models for the data other than independent features are needed for high-dimensional analysis.
\end{abstract}

%% file: sections/intro_v3.tex
\section{Introduction}

Analysis of kernel methods have seen a resurgence after \cite{jacot2018neural} showed an equivalence of wide neural networks, trained with gradient descent, with the so-called \textit{neural tangent kernel} (NTK). 
Contemporaneously, there has been growing interest
in high-dimensional asymptotic analyses of machine learning methods in a regime where the number of input samples $n$ and number of input features $p$ grow proportionally as
\begin{equation} \label{eq:proportional_asymptotics}
    p/n \rightarrow \beta,
\end{equation}
for some $\beta > 0$ and the data follow some random distribution.  This regime often enables remarkably precise predictions on the behavior
of complex algorithms (see, e.g.~
\citep{krzakala2012probabilistic},
and the references below).

In this work, we study kernel estimators of the form:
\begin{equation} \label{eq:ykernel}
    \yhat(x) = \sum_{i=1}^n K(x,x_i)\alpha_i,
\end{equation}
where $\{\alpha_i\}_{i=1}^n$
are weights learned from training samples $\{(x_i,y_i)\}_{i=1}^n$, and $K(\cdot,\cdot)$ is a kernel.  
We consider the training of such kernel models in an asymptotic random regime similar 
in form to several other high-dimensional analyses:

\textbf{Proportional, uniform large scale limit:}  Consider 
a sequence of problems indexed by the number of data samples $n$ satisfying the following assumptions:
\begin{enumerate}[label=A\arabic*, itemsep=0pt, topsep=0pt]
    \item (Uniform data) \label{as:covariance} Training features are generated as 
    $x_i = \Sigma_x^{1/2}z_i \in\Real^p$ where $z_i\in \Real^p$ has i.i.d. entries with $\Exp z_i = 0$, $\Exp |z_i|^2 = 1$, and $\Exp |z_i|^{5+\varepsilon}<\infty$ for some $\varepsilon >0$. 
    A test sample, $x_{\rm ts} = \Sigma_x^{1/2} z_{\rm ts}$,
    is generated similarly.
    Further, the covariance matrix
 $\Sigma_x$ is positive definite with $\|\Sigma_x\|_2 = \mathcal{O}(1)$, and\\ $\tau:=\lim_{p\rightarrow \infty}\tr(\Sigma_x)/p<\infty$.
 \item (Proportional asymptotics) Number of samples $n$ and number of input features $p$ scale as \eqref{eq:proportional_asymptotics}.

 \item (Kernel) \label{as:continuity} The kernel function is of the form
 \begin{equation} \label{eq:kernel_form}
    K(x_i,x_j) = g\left(\frac{\tnorm{\xbf_i}^2}{p}, \frac{\inner{\xbf_i, \xbf_j}}{p}, \frac{\tnorm{\xbf_j}^2}{p}\right)
\end{equation}
 where $g$ is $C^1$ around $(\tau, \tau, \tau)$, $C^3$ around $(\tau, 0, \tau)$.
\end{enumerate}

Under these assumptions we show that:
\begin{center}
    \mbox{
\textit{Kernel regression offers no gain over linear models.}}
\end{center}

Our result does not disregard kernel methods (or neural networks)
as a whole, but serves as a caution regarding the
proportional uniform large scale limit model 
while examining the asymptotic properties of kernels. A result of this nature regarding the high-dimensional degeneracy of two layer neural networks has been studied in \cite{hu2020surprising}.

\subsection{Summary of Contributions}
To be  precise,
we show three surprising results concerning kernel regression
in the proportional, uniform large scale limit:
\begin{enumerate}[label=\arabic*., leftmargin=4mm, topsep=1mm]
    \item First, we show  
    kernel models only learn linear relations between the covariates $x$ and the response $y$ in this regime. 
    Consequently, kernel models (including neural networks in the kernel regime) have no benefit over linear models in this regime. 
    \item Our second result considers the training dynamics of the kernel and linear models. We show that under gradient descent, in the high dimensional setting, dynamics of the kernel model and a scaled linear model are equivalent throughout training.
    \item Finally, we consider the case where the true data is generated from a kernel model with some unknown parameters. In this case, the relation between $x$ and $y$ can be highly nonlinear.  An example of such a model is that $y$ is generated from $x$ via a neural network with random, unknown parameters.  In this case, we show that in the high-dimensional limit, the linear networks provide the minimum generalization error. That is, again, nonlinear kernel methods provide no benefit and training a wide neural network would result in a linear model.
\end{enumerate}

The main take-away of this paper is that under certain data distribution assumptions that are widely used in theoretical papers, a large class of kernel methods, including fully connected neural networks (and residual architectures with fully connected blocks) in kernel regime, can only learn linear functions. Therefore, in order to theoretically understand the benefits that they provide over linear models, more complex data distributions should be considered. Informally, if $x \in \Real^p$ covers this space in every direction (not necessarily isotropically), and the number of samples grows only linearly in the dimension of this space, many kernels can only see linear relationships between the covariates and the response. In other words, we argue that if we seek high-dimensional models for analyzing performance of neural networks, other distributional assumptions will be needed.

The proofs of our results rely on a generalization of Theorem 2.1 and 2.2 of \citep{el2010spectrum}. This generalization might be of independent interest for other works.

\subsection{Prior work:}
High-dimensional analyses in the proportional asymptotics regime similar to assumptions~ \ref{as:covariance} to \ref{as:continuity} have been widely-used in statistical physics and random matrix-based analyses of inference algorithms~\citep{zdeborova2016statistical}.  The high-dimensional framework has yielded powerful results in a wide range of applications such as estimation error in linear inverse problems \citep{donoho2009message, bayati2011dynamics, krzakala2012probabilistic,rangan2019vector, hastie2019surprises}, convolutional inverse problems \citep{sahraee2021asymptotics}, dynamics of deep linear networks \citep{saxe2013exact}, matrix factorization \citep{kabashima2016phase}, binary classification \citep{taheri2020sharp, kini2020analytic}, inverse problems with deep priors \citep{gabrie2019entropy,pandit2019inference, pandit2020matrix}, generalization error in linear and generalized linear models \citep{gerace2020generalisation,emami2020generalization, loureiro2021learning, gerbelot2020asymptotic}, random features \citep{ascoli20a}, and for choosing the optimal objective function for regression \citep{bean2013optimal, advani2016statistical} to name a few. 
Our result that, under a similar set of assumptions, kernel regression degenerates to linear models is thus somewhat surprising.  

That being said, the result is not entirely new.
Several authors have suggested that
high-dimensional data modeled with 
i.i.d.\ covariates are inadequate
\citep{goldt2020modeling,mossel2016deep}.
The results in this paper can thus be seen
as attempting to describe the limitations precisely.

In this regard, the work is closest to \citep{hu2020surprising}.  The work
\citep{hu2020surprising}
proves that for a two-layer fully-connected neural network, the training dynamics are equivalent to a linear model in inputs. They provide asymptotic rates for convergence in the early stages of training ($t< O(p \log p)$). Our result, however, considers a much larger class of kernels and is not limited to the NTK. In addition, we consider the dynamics 
throughout the training including the limit.

The generalization of kernel ridgeless regression is also discussed in this setting  in \citep{liang2020just}. The connections to double descent with explicit regularization has been analyzed in \citep{liu2021kernel}. The authors in \citep{dobriban2018high}, characterize the limiting predictive risk for ridge regression and regularized discriminant analysis. \citep{cui2021generalization} provides the error rates for KRR in the noisy case, and the generalization error in learning with random features with kernel approximation has been discussed in   \citep{liu2020random}. A comparison between neural networks and kernel methods for Gaussian mixture classification is also is provided in \citep{refinetti2021classifying}.

The kernel approximation of the over-parameterized neural networks does not limit their performance in practical applications. In fact, these networks have surprisingly shown to generalize well \citep{neyshabur2017exploring, zhang2021understanding, belkin2018understand}. Of course, in the non-asymptotic regime, these models also have very large capacity \citep{bartlett2017spectrally}. While this high capacity leads to learning complex functions, it is not always the case for the trained networks, and large models might still advocate for learning simpler functions. Works such as \citep{NEURIPS2019_b432f34c, hu2020surprising} show that this simplicity can come from the implicit regularization induced by the training algorithms such as gradient descent for early-time dynamics. In this work, however, we show that in the high dimensional limit, this simplicity can be a result of the uniformity of input distribution over the space. In fact, we show that in this regime, kernel methods are no better than linear models.

%% file: sections/main_resutls.tex
\section{Kernel Methods Learn Linear Models}\label{sec:main_results}
In this section we show the first result of this paper:  in the proportional, uniform high-dimensional regime, 
fitting kernel models is equivalent to fitting a regularized least squares model with appropriate regularization parameters. A short review of reproducing kernel Hilbert spaces (RKHS) and kernel regression can be found in Appendix \ref{sec:kernel_regression}.

Suppose we have $n$ data points $(x_i,y_i)$, $i=1,\ldots,n$ with $x_i\in\Real^p$, and an RKHS $\mcH$ corresponding to the kernel $K(\cdot, \cdot)$.

Consider two models fitted to this data:
\begin{enumerate}[label= \arabic*.]
    \item  Kernel ridge regression model $\wh f_{\krr}$ which solves
\begin{equation}
    \wh f_\krr = \argmin_{f\in\mcH} \sum_{i=1}^n (y_i - f(x_i))^2 + \lambda \|f\|_{\mcH}^2,\label{eq:kernel_ridge}
\end{equation}
where $\|f\|_{\mcH}=\sqrt{\innerH{f,f}}$ is the Hilbert norm of the function.
\item  Linear model $
\wh f_{\lin}(x) = \wh w\tran x + \wh b$ fitted by solving the $\ell_2$-regularized least squares problem:
\end{enumerate}
\vspace{-5mm}
\begin{align}
    (\widehat{w},\widehat{b}) =& \argmin_{w,b} J(w,b), \label{eq:linear_model} \\
    J(w,b) :=& \sum_{i=1}^n (y_i - w\tran x_i - b)^2  + \lambda_1 |b|^2 + \lambda_2 \|w\|^2.
\end{align}
The problem in \eqref{eq:kernel_ridge} is an optimization over a function space. By parameterizing $f_\krr$ as $f_\krr(x)=\inner{\phi(x), \theta}$ we can find the optimal function by solving
\begin{equation}
    \widehat{\theta} = \argmin_{\theta\in \mc H} \sum_{i=1}^n \mc (\left(y_i - \inner{\phi(x_i), \theta}\right)^2 + \lambda \|\theta\|_{L^2}^2.
\end{equation}
By the representer theorem \citep{scholkopf2001generalized}, the optimal function in \eqref{eq:kernel_ridge} also has the form
\begin{equation}
    \fkhat(x) = \sum_{i=1}^n K(x, x_i)\alpha_i,
\end{equation}
where $\alpha$ solves
\begin{equation}
    \min_{\alpha}\ \sum_{i=1}^n \mc (y_i - \Kbf_i \alpha)^2 + \lambda \alpha\tran \Kbf \alpha,
\end{equation}
where $\Kbf\in \Real^{n\times n}$ with $\Kbf_{ij} = K(x_i, x_j)$ is the kernel matrix and $\Kbf_i$ is its $i^{\rm th}$ row.

To state the result we need to define the following constants related to the kernel and its associated function $g$ from assumption \ref{as:continuity}
\begin{subequations}\label{eq:c_def}
\begin{align}
    c_2 &= g'(\tau, 0, \tau),\label{eq:c2_def}\\
    c_0 &= g(\tau, \tau, \tau) - g(\tau, 0, \tau) -c_2\frac{\trace{\Sigma_p}}{p},\label{eq:c0_def}\\
    c_1 &= g(\tau, 0, \tau) +g''(\tau, 0, \tau)\frac{\trace{\Sigma_p^2}}{2p^2}.\label{eq:c1_def}
\end{align}
\end{subequations}
where $g'$ and $g''$ are partial derivatives of $g$ in the second argument.

Our first result shows that with an appropriate choice of $(\lambda, \lambda_1, \lambda_2)$ the two models $\wh f_\krr$ and $\wh f_{\lin}$ are in fact equivalent.

\begin{theorem}\label{thm:equivalence_linear_kernel}
Under Assumptions (\ref{as:covariance}-\ref{as:continuity}), if we use the same data to train $\wh f_\krr$ and $\wh f_{\lin}$ with
\begin{equation}\label{eq:Lambda_eqv}
    \lambda_1 = \frac{c_0 + \lambda}{c_1}, \quad \lambda_2 = \frac{p(c_0 + \lambda)}{c_2},
\end{equation}
where the constants $c_0, c_1$, and $c_2$ are defined in equations \eqref{eq:c_def},\noeqref{eq:c2_def}\noeqref{eq:c0_def}\noeqref{eq:c1_def}  then at a test sample, $x_{\rm ts}$ drawn from the same distribution as the training samples,
\begin{equation*}
    \lim_{n,p\rightarrow \infty} |\fhat_{\lin}(x_{\rm ts}) - \wh f_\krr(x_{\rm ts})| \eqp 0.
\end{equation*}
\end{theorem}
\begin{proof}
See Appendix \ref{sec:proof_of_ElK}.
\end{proof}

\begin{remark}
Note that the result in Theorem \ref{thm:equivalence_linear_kernel} does not imply that the linear model and the kernel model are equal in probability for all the points in the domain of these functions in the proportional uniform regime, but rather over a random test point as given by assumption \ref{as:covariance}. However this suffices for understanding the generalization properties of these functions.
\end{remark}

\begin{remark}
Since convergence in probability implies convergence in distribution, we also have that the generalization error of $\wh f_\krr$ is the same as that of $\wh f_\lin$ for any bounded continuous metric.
\end{remark}

\begin{remark}
Theorem \ref{thm:equivalence_linear_kernel} states a convergence in probability for a single test point. This holds for $n_\ts$ test samples so long as $n_\ts$ grows at most linearly in the number of training samples, i.e. $n_\ts = \mc O(n_\tr),$ and the outputs of kernel model and the linear model would be equal in probability over all these test samples.
\end{remark}

\section{Linear Dynamics of Kernel Models}
Our next result shows that if a kernel ridge regression is solved using gradient descent, every intermediate estimator during training has an equivalent linear model.

Consider a kernel model that is parameterized as $\wh f(x) = \inner{\phi(x), \widehat{\theta}}$ (where $\phi(x)=K(x,\cdot)$ is the feature map) that is trained by regularized empirical risk minimization:
\begin{align}
        \widehat{\theta} = \argmin_{\theta} \sum_{i=1}^n (y_i - \inner{\phi(x_i),\theta})^2 + \lambda \|\theta\|_{L^2}^2.
\end{align}
The gradient descent iterates for this problem are,
\begin{equation}
    \theta^{t+1} = (I -\eta ((\phi(X_\tr)\tran \phi(X_\tr) + \lambda I))\theta^t + \eta \phi(X_\tr)\tran y_\tr.
\end{equation}
with $\theta^0=0$. Here, $\phi(X_\tr)$ is a matrix with $\phi(x_i)$ as its $i$th row and $\eta$ is the learning rate. Similarly, consider a scaled linear model $f_{\rm lin}(x) = \gamma_2 w\tran x + \gamma_1 b$ learned by optimizing, via gradient descent, the regularized squared loss:
\begin{align}
    \MoveEqLeft 
    \widehat{w}, \widehat{b} = \argmin_{w, b} \|y_\tr - \gamma_2 X_\tr w - \gamma_1 b1\|^2  \nonumber \\
    &+ \lambda_1 |\gamma_1 b|^2 + \lambda_2 \|\gamma_2 w\|^2.
\end{align}
The parameters are initialized zero in gradient descent. Observe that this optimization problem is equivalent to the one in \eqref{eq:linear_model} as we have only made a change of variables $w$ to $\gamma_2 w$ and $b$ to $\gamma_1 b$, i.e.\ they learn the same model. The scalings are introduced to make the training dynamics of the linear model and the kernel model the same. Let the parameter of the kernel model after $t$ steps of gradient descent be $\theta^t$ and define $\fkhat^t(x) = \inner{\phi(x), \theta^t}$. Similarly, let the parameters of the linear model after $t$ step of gradient descent be $(w^t, b^t)$ and define $\widehat{f}^t_{\rm lin}(x) = \gamma_2 {w^t}\tran x + \gamma_1 b^t$. Then we have the following result.
\begin{theorem}\label{thm:equivalence_GD}
If $\gamma_1 = \sqrt{c_1}, \gamma_2=\sqrt{c_2/p}$ and $\lambda_1, \lambda_2$ are given by equation \eqref{eq:Lambda_eqv}, then for any step $t\geq 0$ of gradient descent (initialized at zero) and any test sample drawn from the same distribution as the training data we have
\begin{equation}
     {\fkhat}^t(x_\ts) \eqp \widehat{f}_{\rm lin}^t(x_\ts).
\end{equation}
\end{theorem}
\begin{proof}
The proof can be found in Appendix \ref{sec:proof_of_equivalence_GD}.
\end{proof}

\begin{remark}
Theorem \ref{thm:equivalence_GD} provides an insight into the training dynamics of kernel models in the proportional uniform regime. This could potentially have implications regarding the Kernel-SVM solution in this regime, following the work of \cite{muthukumar2021classification}.
\end{remark}

\section{Optimality of Linear Models}
Our last result shows that in the proportional uniform large scale limit, if the true model has a Gaussian process prior with a kernel that satisfies assumption \ref{as:continuity}, then linear models are in fact optimal, even though the true underlying relationship between the covariates and the responses could be highly nonlinear. See Appendix \ref{sec:Gaussian_process_regression} for a review of Gaussian process regression.

Assume that we are given $n$ training samples $(x_i, y_i)$ 
\begin{equation}
    y_i = f^*(x_i) + \xi_i, \quad \xi_i \stackrel{i.i.d.}{\sim}\mcN(0,\sigma^2),\label{eq:Gaussian_model}
\end{equation}
and the function $f^*$ is a zero mean Gaussian process with covariance kernel $K(\cdot, \cdot)$. An example occurs in the so-called student-teacher set-up
of \citep{gardner1989three,aubin2019committee} where
the unknown function is of the form
\begin{equation} \label{eq:nntrue}
    f(x) = g(x,\theta),
\end{equation}
and $g(x,\theta)$ is a neural network
with unknown parameters $\theta$.
If the network
has infinitely wide hidden layers and 
the unknown 
parameters $\theta$ are generated
with randomly with i.i.d.\ Gaussian
coefficients with the appropriate scaling,
the unknown function $f(x)$ in
\eqref{eq:nntrue} becomes asymptotically
a Gaussian process \cite{neal2012bayesian, lee2017deep, matthews2018gaussian, daniely2016toward}.

Now assume that we are given a test sample from the same model $(x_\ts, y_\ts)$ and we are interested in estimating $y_\ts$. It is well known (see Appendix \ref{sec:Gaussian_process_regression}) that the Bayes optimal estimator with respect to squared error in this case is
\begin{equation}
    \fhat_{\rm opt}(x_\ts) = K(x_\ts,X_\tr)(\Kbf + \sigma^2 I)^{-1} y_\tr,\label{eq:kernel_model_opt}
\end{equation}
and its Bayes risk is
\begin{equation}
    \mc E_{\opt}(x_\ts) = \sigma^2 + K(x_\ts, x_\ts) - K(x_\ts,X_\tr)(\Kbf + \sigma^2 I)^{-1}K(X_\tr,x_\ts).
\end{equation}
Next consider a linear model $\fhat_{\rm lin}(x) = \widehat{w}\tran x + \widehat{b}$ fitted by solving the regularized least squares problem in \eqref{eq:linear_model}. Define the square error risk of this model as
\begin{equation}
    \mc E_{\lin}(x_\ts) = \Exp [(y_\ts - \fhat_{\lin}(x_\ts))^2|x_\ts, X_\tr, y_\tr],
\end{equation}
where the expectation is with respect to the randomness in $f$ as well as the noise $\xi_\ts$.

\begin{theorem}\label{thm:optimality_of_linear} 
Under assumptions (\ref{as:covariance}-\ref{as:continuity}) and the Gaussian data model \eqref{eq:Gaussian_model} if the linear model $\wh f_\lin$ in equation \eqref{eq:linear_model} is trained with regularization parameters
\begin{equation}
    \lambda_1 = \frac{c_0 + \sigma^2}{c_1}, \quad \lambda_2 = \frac{p(c_0 + \sigma^2)}{c_2}.\label{eq:optimal_reg}
\end{equation}
where $c_0, c_1$ and $c_2$ are defined in Proposition \ref{thm:generalization_of_ELK_result}, then $\wh f_\lin$ achieves the Bayes optimal risk for any test sample drawn from the same distribution as training data, 
\begin{equation} \label{eq:Emselslim}
    \lim_{n\rightarrow \infty}  \left| \Ec_{\lin}(x_\ts)
    - \Ec_{\opt}(x_\ts) \right| \eqp 0.
\end{equation}
\end{theorem}
\begin{proof}
The result of Theorem \ref{thm:equivalence_linear_kernel} shows that with the choice of regularization parameters in \eqref{eq:optimal_reg}, the linear model and the kernel model in \eqref{eq:kernel_model_opt} are equivalent
\begin{equation*}
    \lim_{n\rightarrow\infty} \fhat_{\lin}(x_\ts) \eqp \fhat_{\opt}(x_\ts).
\end{equation*}
The result then immediately follows as the kernel model is Bayes optimal for squared error.
\end{proof}

It is important to contrast this result with \citep{goldt2020dynamics} and
\citep{aubin2019committee}.  The
works  \citep{aubin2019committee,goldt2020dynamics}
consider exactly the case
where the true function is of the
form \eqref{eq:nntrue} where $g(x,\theta)$
is a neural network with Gaussian i.i.d.\ 
parameters.  However, in their analyses,
the number of hidden units in both
the true and trained network are 
\emph{fixed}
while the dimension of $x$ and number of
samples grow with proportional scaling.
With a fixed number of hidden units,
the true function is \emph{not} a Gaussian process, and the model class is not
a simple kernel estimator -- hence, 
our results do not apply.  Interestingly,
in this case, the results of 
\citep{aubin2019committee,goldt2020dynamics} show that 
nonlinear models
can significantly out-perform linear models.    Hence, very wide neural networks can underperform networks with smaller numbers of hidden units.
It is an open question as to which scaling of the number of hidden units, 
number of samples, and dimension
yield degenerate results.

%% file: sections/proof_sketch.tex
\section{Sketch of Proofs}
Here we provide the main ideas behind the proofs of our main theorems. The details of the proof of Theorem \ref{thm:equivalence_linear_kernel} can be found in Appendix \ref{sec:proof_of_equivalence}. Proof of Theorem \ref{thm:equivalence_GD} can be found in Appendix \ref{sec:proof_of_equivalence_GD}.
\subsection{Degeneracy of empirical kernel matrices}

Our first result extends Theorems 2.1 and 2.2 of \cite{el2010spectrum} and may be of independent interest to the reader.

\begin{proposition}\label{thm:generalization_of_ELK_result}
If $\Kbf$ is a $n\times n$ kernel matrix with entries \eqref{eq:kernel_form} such that assumptions (\ref{as:covariance}-\ref{as:continuity}) hold, then
\begin{equation*}
    \lim_{p\rightarrow \infty}\tnorm{\Kbf - \Mbf} \eqp 0,
\end{equation*}
where $\Mbf = c_0 \Ibf + c_1\ones \ones\tran + c_2 X X\tran$ where $c_0, c_1, c_2$ are defined in equation \eqref{eq:c_def} and $X\in\Real^{n\times p}$ is the design matrix with samples $\{x_i\}$ as rows.
\end{proposition}
\begin{proof}
See appendix \ref{sec:proof_of_ElK}.
\end{proof}
\cite{el2010spectrum} present this result for kernels of the form $h(\inner{x_i, x_j})$ or $h(\|x_i-x_j\|_2^2)$. Importantly, the NTK has a form that is neither $h(\inner{x_i, x_j})$ or $h(\|x_i-x_j\|_2^2)$, but in fact of the form in equation \eqref{eq:kernel_form}, whereby Proposition \ref{thm:generalization_of_ELK_result} provides new insights into the behaviour of empirical kernel matrices of the NTK for a large class of architectures.

\subsection{Equivalence of Kernel and Linear Models}
Proposition \ref{thm:generalization_of_ELK_result} is the main tool we use to show that kernel methods and linear methods are equivalent in the proportional, uniform large scale limit. 

The model learned by the kernel ridge regression in equation \eqref{eq:kernel_ridge} can be written as
\begin{equation}
    \wh f_\krr(x) = K(x, X_\tr)(K(X_\tr, X_\tr) + \lambda I)^{-1} y_\tr.\label{eq:kernel_estimator}
\end{equation}
Next, since the optimization in \eqref{eq:linear_model} is a quadratic problem it has a closed form solution.
\begin{proposition}
The linear estimator in Model 2 has the following form
\begin{equation}
    \wh f_{\rm lin}(x) =
    \left(\frac{x\tran X_\tr\tran}{\lambda_2} + \frac{1\tran}{\lambda_1}\right) \left[\frac{X_\tr X_\tr\tran}{\lambda_2} + \frac{11\tran}{\lambda_1} + I_n\right]^{-1}y_\tr.\label{eq:yhatls_solution}
\end{equation}
\end{proposition}
\begin{proof}
Solving the optimization problem in \eqref{eq:linear_model} we get
\begin{align*}
    \left[\begin{matrix}\what\\ \bhat\end{matrix}\right]
    &=  \left[\begin{matrix}X\tran_\tr X_\tr + \lambda_2 I_n & X_\tr\tran 1\\ 1\tran X_\tr & n+\lambda_1\end{matrix}\right]^{-1}
     \left[\begin{matrix}X_\tr\tran\\ 1\tran\end{matrix}\right] y_\tr\\
     & = \left[\begin{matrix}\frac{1}{\lambda_2}X_\tr\tran\\ \frac{1}{\lambda_1}1\tran\end{matrix}\right]\left[\frac{1}{\lambda_2}X_\tr X_\tr\tran + \frac{1}{\lambda_1}1 1\tran + I_n\right]^{-1}y_\tr,\label{eq:solution_linear_model}
\end{align*}
where the last equality follows from a special case of the Woodbury matrix identity (See Lemma \ref{lem:matrix_inversion} in Appendix \ref{app:technical}). 
\end{proof}

Next, we can use Proposition \ref{thm:generalization_of_ELK_result} to show that each of the terms in Equation \eqref{eq:yhatls_solution} converge in probability to the corresponding term in Equation \eqref{eq:kernel_estimator}. This proves Theorem \ref{thm:equivalence_linear_kernel}.

\subsection{Equivalence Throughout Training}
The proof of equivalence of the kernel model and scaled linear model after $t$ steps of gradient descent is very similar. The updates for parameters of the kernel model as well as the parameters of the scaled linear model have linear dynamics (in their respective parameters). By unrolling the gradient update through time, we can write the parameters after $t$ step as a summation over the past time steps. Using this, we can simplify the sums to write the output of the kernel model at time $t$ over a test sample as
\begin{equation}
    \fkhat^t(x_\ts) 
     = K(x_\ts, X_\tr)((K(X_\tr, X_\tr) + \lambda I)^{-1}
    \bigg(I - 
    \big(I -\eta ((K(X_\tr, X_\tr) + \lambda I)\big)^{t}\bigg)y_\tr.
\end{equation}
Similarly, for the linear model at step $t$ we get
\begin{multline}
    f^t_{\rm lin}(x_\ts)
    = (\gamma_2^2x_\ts X_\tr \tran + \gamma_1^2 1\tran)(\gamma_2^2X_\tr X_\tr \tran + \gamma_1^2 1 1\tran +\gamma_2^2\lambda_2 I)\inv\\
    \left(I - \left(I - \eta (\gamma_2^2X_\tr X_\tr \tran + \gamma_1^2 1 1\tran +\gamma_2^2\lambda_2 I)\right)^{t}\right)y_\tr.
\end{multline}
Here, we can use Proposition \ref{thm:generalization_of_ELK_result} again to show that all the terms in the linear model converge in probability to the corresponding term in the kernel model, thus proving that the two models are equal in probability for any test sample drawn from the same distribution as the training data over the course of gradient descent.

%% file: sections/experiments.tex
\section{Numerical Experiments}
\subsection{Linearity of Kernel Models for NTK}\label{sec:exp_equivalence_of_linear}

We demonstrate via numerical simulations the predictions made by our results in Theorems \ref{thm:equivalence_linear_kernel}, \ref{thm:equivalence_GD}, \ref{thm:optimality_of_linear}

As shown in \cite{lee2019wide} and \cite{liu2020linearity}, wide fully connected neural networks can be approximated by their first order Taylor expansion throughout the training
\begin{equation*}
    f_{\lin}(x) = f(x; \theta_0) + \inner{\nabla_\theta f(x, \theta_0), \theta - \theta_0},
\end{equation*}
and this approximation becomes exact in the limit that all the hidden dimensions of the neural network go to infinity.
Therefore, training a network $f(x;\theta)$ by minimizing
\begin{equation}
    \widehat{\theta} = \argmin_\theta \sum_{i=1}^n\left(y_i - \big(f(x;\theta) - f(x;\theta_0)\big)\right)^2 + \lambda \|\theta - \theta_0\|_2^2\label{eq:neural_network_train}
\end{equation}
is equivalent (in the limit of wide network) to performing kernel ridge regression in an RKHS with feature map $x\mapsto \nabla_\theta f(x;\theta_0)$ and neural tangent kernel $K_L(x,x')$ as its kernel\footnote{See Appendix \ref{app:NTK} for a brief review of NTK.}. Instead of removing the initial network, one can use a symmetric initialization scheme which makes the output of neural network zero at initialization without changing its NTK \cite{chizat2018lazy, zhang2020type, hu2020surprising}.

A key property of the NTK of fully connected neural networks is that it satisfies assumption \ref{as:continuity} since it has the form in equation \eqref{eq:kernel_form}. Hence, if the input data $x$ satisfies the requirements of this theorem, in the proportional asymptotics regime the NTK should behave like a linear kernel. 
The first and second order derivatives of the kernel function can be obtained by backpropagation through the recursive equations in \eqref{eq:NTK_sigma_rec} and \eqref{eq:NTK_K_rec}.

\begin{figure}[t]
    \centering
    \includegraphics[width=0.55\textwidth]{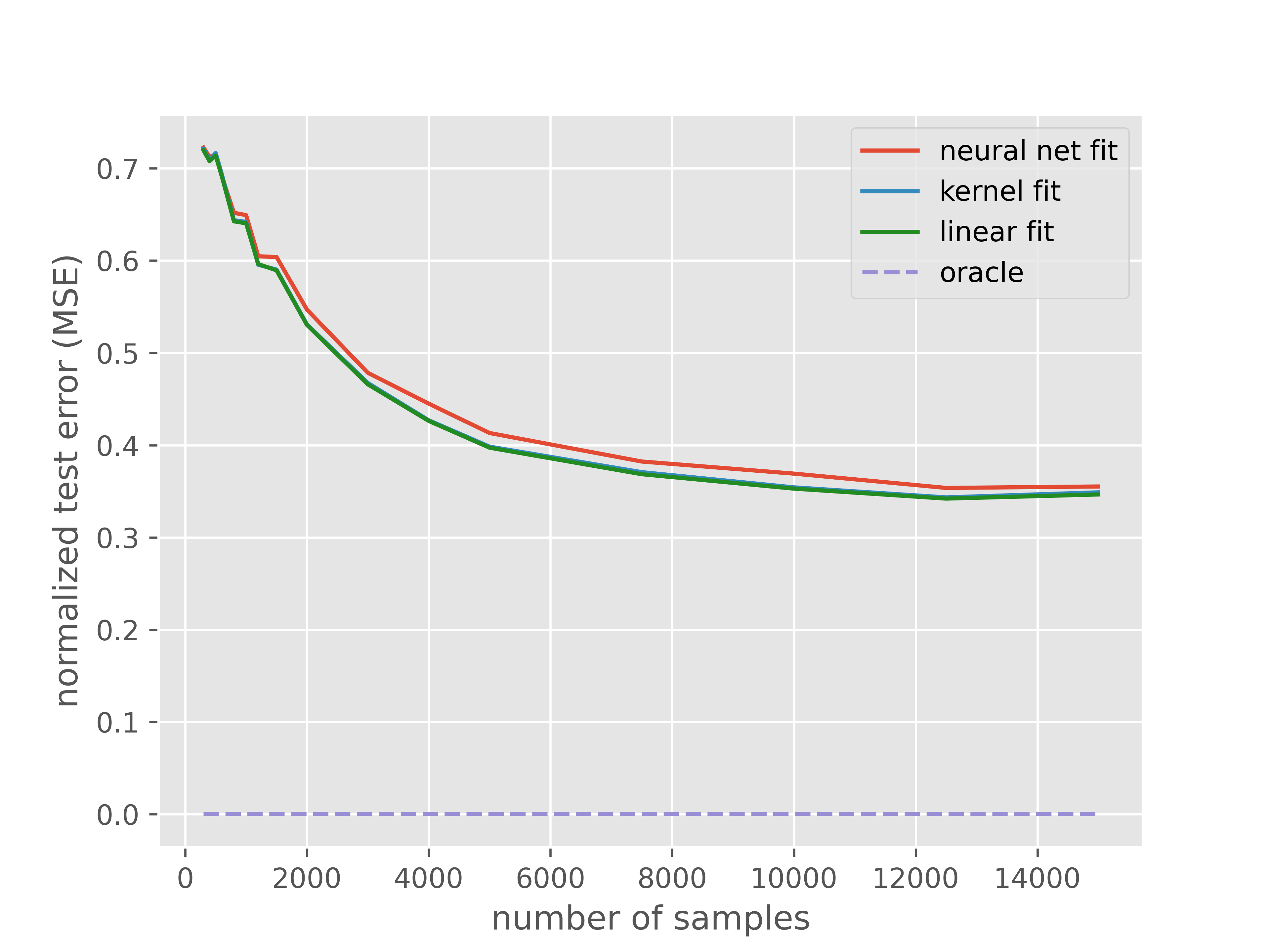}
    \caption{Comparison of test error for three different models: (i) a neural network with a single hidden layer, (ii) NTK of a two layer fully-connected network, and (iii) the linear equivalent model prescribed by Theorem \ref{thm:equivalence_linear_kernel}. The errors of the kernel model and the equivalent linear model match perfectly and neural network follows them very closely. The oracle model is the true model and represents the noise floor. We use $\lambda = 0.005$.}
    \label{fig:nn_results}
\end{figure}

Figure \ref{fig:nn_results} illustrates a setting where kernel models and neural networks in the kernel regime perform no better than appropriately trained linear models. This verifies the main result of this paper -- Theorem \ref{thm:equivalence_linear_kernel}.

We generate training data for $i=1,2,\ldots, n$ as
\begin{equation}
    y_i = f^*(x_i) +\xi_i, \quad x_i\sim \mc N(0, I_{p\times p}),\ \ \xi_i \sim \mcN(0, \sigma^2),
\end{equation}
where $p=1500$ and $\sigma^2 =0.1$ and $f^*$ is a  fully-connected ReLU network with two hidden layers with 100 hidden units each. 

We train 3 models:
\begin{enumerate}[label=(\roman*)]
    \item A fully connected ReLU neural network with a single layer of $20,\!000$ hidden units to fit this data using stochastic gradient descent (SGD) with momentum parameter $0.9$. The initial network is remove from the output as in \eqref{eq:neural_network_train}. 
    \item A kernel model as in equation \eqref{eq:kernel_ridge} corresponding to the NTK
of the model in (i) above. The kernel is evaluated using the recursive formulae given in \eqref{eq:NTK_sigma_rec} and \eqref{eq:NTK_K_rec}.
    \item A linear model as in equation \eqref{eq:linear_model} trained using the regularization parameters prescribed by Theorem \ref{thm:equivalence_linear_kernel}.
\end{enumerate}

We compare the test error for these models, measured as $1-R^2$ over $n_\ts=200$ test samples:
\begin{equation}
    \mc E_\ts = \frac{\sum_{i=1}^{n_\ts}(y_{\ts,i}-\yhat_{\ts, i})^2}{\sum_{i=1}^{n_\ts}(y_{\ts,i})^2},\label{eq:normalized_error}
\end{equation}
We compare the test error for different number of training samples $n$ averaged over 3 runs. 

We can see that the NTK model and the equivalent linear model almost match perfectly for all the values of test samples and the neural network model follows them very closely, matching them for small number of samples. 

There are two main sources of mismatch between the neural network model and the NTK model: first the width of the network while large (20,000) it is still finite, and secondly the training of the neural network model is stopped after 150 epochs, i.e. the neural network trained differs from the optimal neural network. Finally, the oracle model's performance is the noise floor.

\subsection{Equivalence of Kernel and Linear Models Throughout Training}
\begin{figure}[t]
    \centering
    \includegraphics[width=0.45\textwidth]{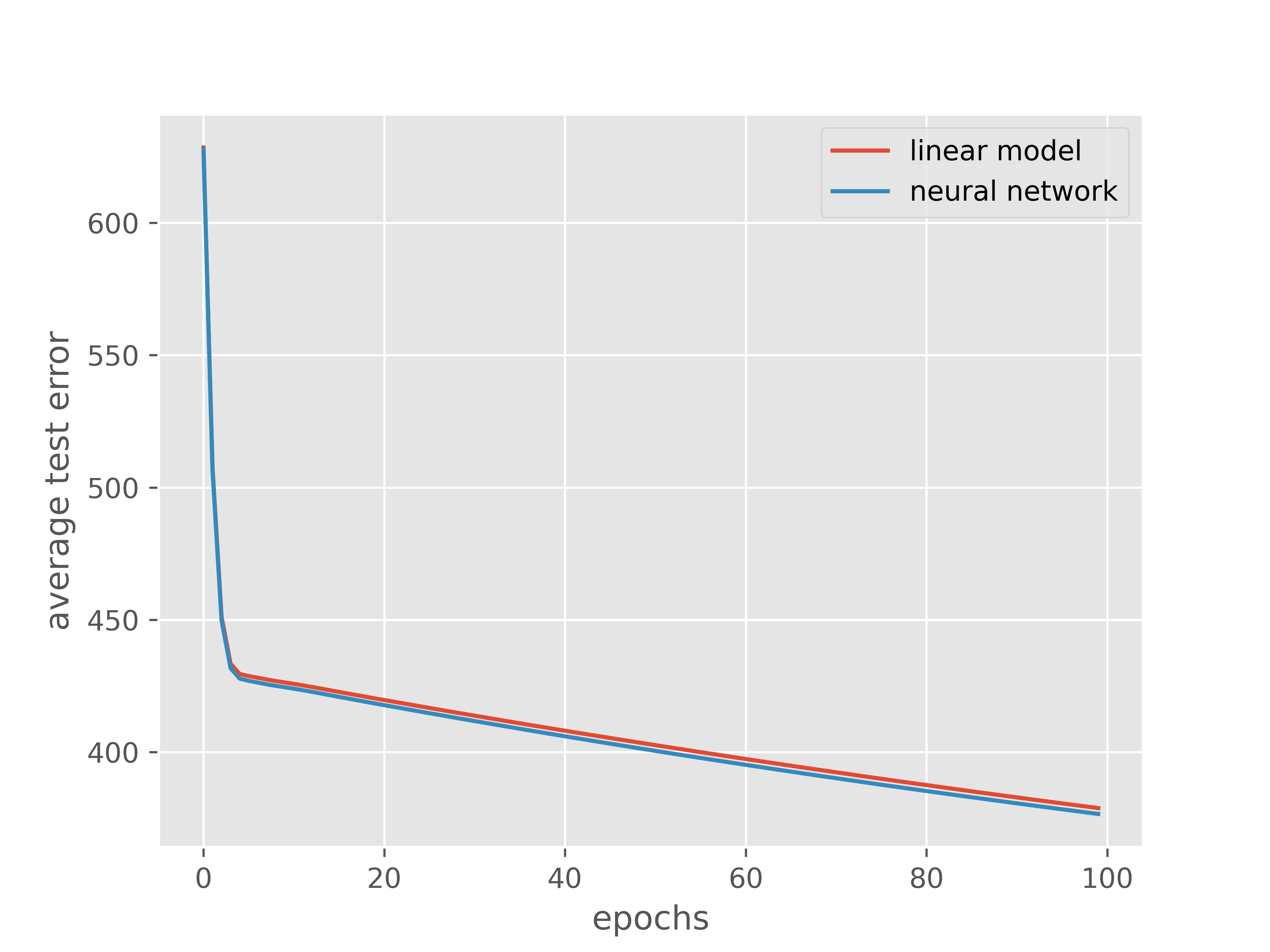}
    \caption{Equivalence of test error of scaled linear model and the neural network vs. epochs of gradient descent.}
    \label{fig:equivalence_GD}
\end{figure}

Next, we verify Theorem \ref{thm:equivalence_GD} by showing that the test error of the scaled linear model and neural network match for all the steps of gradient descent. The setting is the same as in Section \ref{sec:exp_equivalence_of_linear}. We generate data using a random neural network with two hidden layers of $100$ units each and train a neural network with a single hidden layer of 10,000 units as well as the scaled linear model using gradient descent. We plot the the error of each of the models over the test data throughout the training. We train each model for 100 epochs. Figure \ref{fig:equivalence_GD} shows that the two models have the same test error over the course of training.

\subsection{Optimality of Linear Models}\label{sec:exp_optimality_of_linear}
A polynomial kernels of degree $d$ has the following form
\begin{equation}
    K(x, x') = (\inner{x, x'}/p + c)^d,
\end{equation}
where $x, x'\in \Real^p$ and $c\geq 0$ is a constant that adjusts the influence of higher degree terms and lower degree terms. In this examples, we samples test and train samples from the following model
\begin{equation}
    x\sim\mcN(0,I_{p\times p}), \quad y= f(x) +\xi, \quad \xi\sim\mcN(0,\sigma^2),
\end{equation}
where $f$ is a Gaussian process with covariance kernel being a polynomial kernel. We use $c=0.1, d=2$ for the polynomial kernel and set $\sigma^2=0.1$, $p=2,000$. We generate $n_\tr$ samples and train the kernel model and the equivalent linear model and estimate the normalized mean squared error of the estimator by averaging the normalized error over $n_\ts=500$ test samples. We use $\lambda=\sigma^2=0.1$ as the regularization parameter which makes the kernel estimator Bayes optimal (with respect to squared error). The results are averaged over 5 runs. 
\begin{figure}[t]
    \centering
    \includegraphics[width=0.55\textwidth]{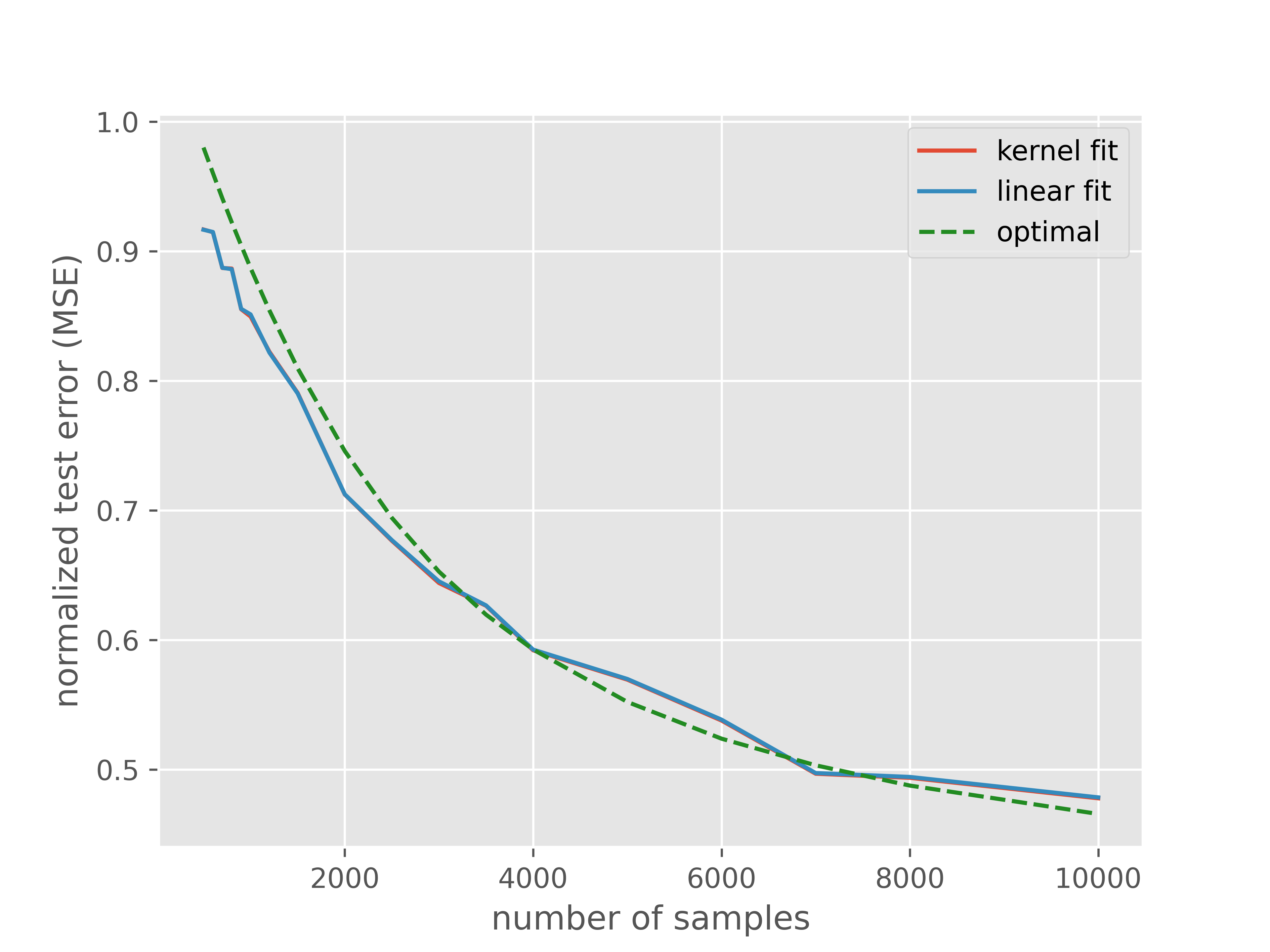}
    \caption{Normalized errors vs.\ number of training samples for a kernel model and the equivalent linear model for a data generated from a Gaussian process. The curves for the kernel and linear fit match almost perfectly. The dashed line corresponds to the theoretical optimal error given in equation \eqref{eq:kernel_model_opt}. 
    \label{fig:Gaussian}}
\end{figure}
The results are shown in Figure \ref{fig:Gaussian} where normalized errors (defined in equation \eqref{eq:normalized_error}) are plotted against the number of training samples. The dashed line corresponds to optimal error curve obtained from Equation \eqref{eq:kernel_model_opt}. The generalization errors for the linear model and the kernel model match which confirm Theorem \ref{thm:equivalence_linear_kernel} and as Theorem \ref{thm:optimality_of_linear} proves both of the curves are very close to the optimal error curve. This figure verifies that the optimal estimator is indeed linear.

\subsection{Counterexample: Beyond the Proportional Uniform Regime}

\begin{figure}[t]
    \centering
    \includegraphics[width=0.55\textwidth]{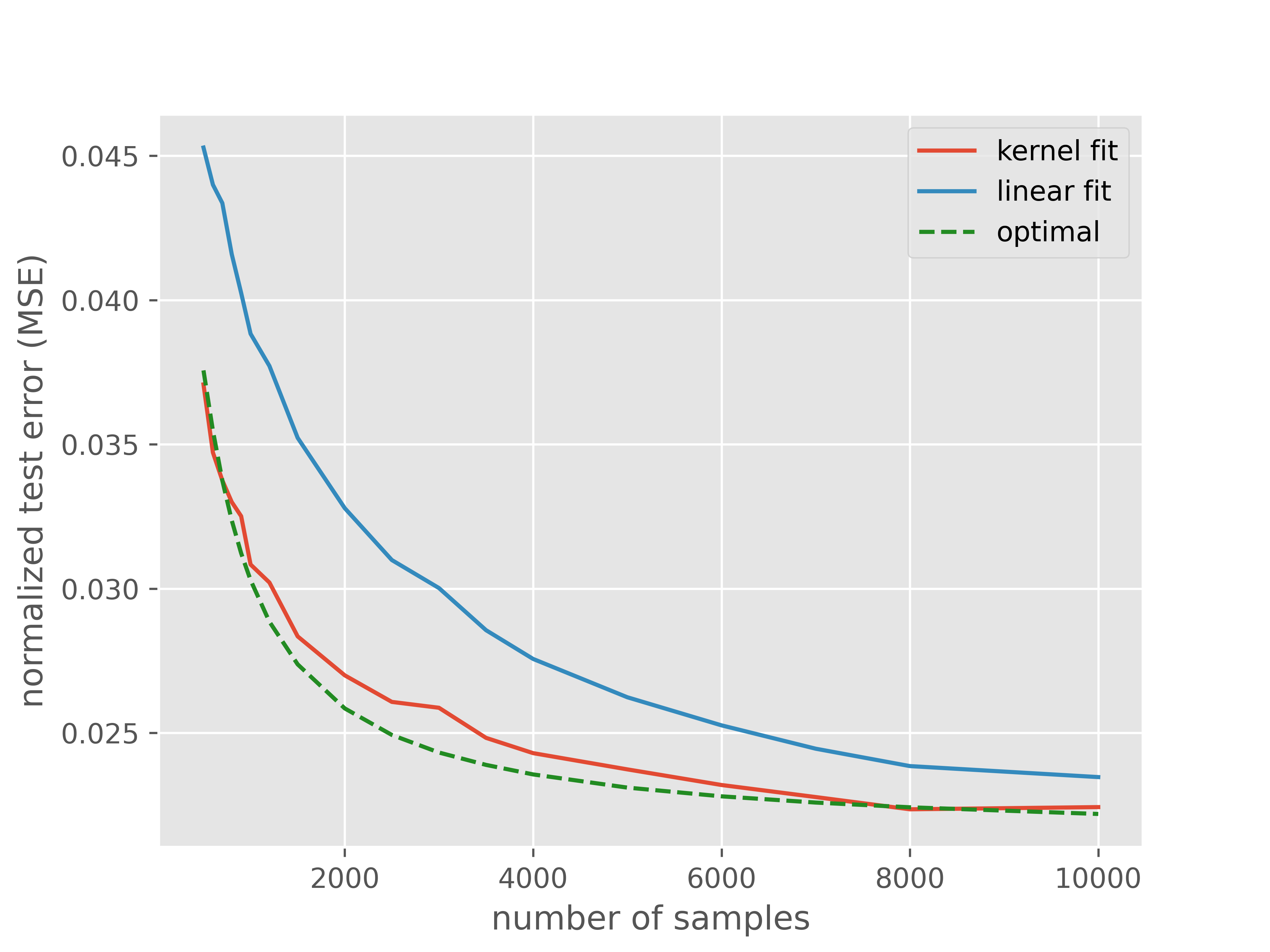}
    \caption{If the assumptions A1-A3 are not satisfied, the kernel model and linear model are not equivalent.}
    \label{fig:mixture}
\end{figure}

Our results should not be misconstrued as ineffectiveness of kernel methods or neural networks. The equivalence of kernel models and linear models holds in the proportional uniform data regime. However kernel models and neural networks outperform linear models when we deviate from this regime, as demonstrated in Figure \ref{fig:mixture}.

This observation is closer to real-world experiences of the machine learning community, which perhaps suggests that the assumptions A1-A3 are unrealistic for understanding high dimensional phenomena relating large datasets and high dimensional models. 

We consider Gaussian process regression as in Section \ref{sec:exp_optimality_of_linear}, but the input variables $x$ are generated from a mixture of two zero mean Gaussians with low-rank covariances, which clearly violates assumption \ref{as:covariance}. The probability of each mixture component is set $1/2$. We use $p=2000$ and set rank of covariance of each component to $r=200$. 
The covariance of each component $c=1,2$ is generated as
\begin{equation}
    \Sigma_c = S_c S_c\tran, \quad S_c\in \Real^{p\times r}, [{S_c}]_{ij}\stackrel{i.i.d.}{\sim}\mcN(0, 1/\sqrt{p}).
\end{equation}
Under this model, the resulting covariance matrix of the data would be 
\begin{equation}
    \Sigma_x = \tfrac{1}{2}\Sigma_1 + \tfrac{1}{2}\Sigma_2,
\end{equation}
which would have rank $2r$ almost surely. In other words, the data only spans a subspace of dimension $400$ of the $2000$-dimensional space. 

Figure \ref{fig:mixture} shows that the kernel model which is the optimal estimator has a generalization error very close to the expected optimal error, whereas the linear model performs worse. The linear approximation $\Mbf$ of the true kernel matrix $\Kbf$ is inaccurate when we deviate from the proportional uniform data regime.

%% file: sections/discussion.tex
\section{Conclusions}

This paper, of course,
\textit{does not} contest the power of neural networks or kernel models
relative to linear models.
In a tremendous range of practical applications, nonlinear models outperform
linear models.  The results should
interpreted as a limitations 
of Assumptions \ref{as:covariance}-\ref{as:continuity}
as a model for high-dimensional data.
While this proportional high-dimensional
regime has been incredibly successful
in explaining complex behavior of many other ML estimators, it provides
degenerate results for kernel models
and neural networks that operate in the kernel regime.

As mentioned above, the intuition is that
when the data samples are generated
as $x = \Sigma_x^{1/2} z$ where $z$ has
i.i.d.\  components and $\Sigma_x$
is positive definite, the data $x$
uniformly covers the space $\Real^p$.
When the number of samples $n$
only scales linearly with $p$,
it is impossible to learn models more
complex than linear models. 

This limitation suggests that more 
complex models for the generated data
will be needed if the high-dimensional
asymptotics of kernel methods are
to be understood.

%% file: sections/appendix.tex
\appendix

\onecolumn
\section*{Appendix}
\input{sections/model}

\input{sections/proofs}

%% file: sections/model.tex
\section{Preliminaries}
We present a short overview of reproducing kernel Hilbert spaces, Gaussian regression, and neural tangent kernels which are used throughout the paper in this appendix.

\subsection{Kernel Regression}\label{sec:kernel_regression}

In kernel regression, the estimator $\widehat{y}(x)$ is a function that belongs to a reproducing kernel Hilbert space (RKHS). A kernel $K:\Real^p\times \Real^p\mapsto \Real$ that is an inner product in a possibly infinite dimensional space $\mc H$ called the \emph{feature space}, i.e.\ $K(x, x') =\innerH{\phi(x), \phi(x')}$ where $\phi:\Real^p\rightarrow\mc H$ is called the feature map. With this feature map, the functions in the RKHS are of the form $f(x) = \innerL{\phi(x), \theta}$ which is a nonlinear function in $x$ but linear in the parameters $\theta$.
In this work, we consider kernels of the form in equation \eqref{eq:kernel_form},
which includes inner product kernels as well as shift-invariant kernels. Many commonly used kernels such as RBF kernels, polynomial kernels, as well as the neural tangent kernel are of this form.

In kernel methods, the estimator is often learned via a regularized ERM
\begin{equation}
    \fkhat = \argmin_{f\in\mcH} \sum_{i=1}^n \mc L\left(y_i , f(x_i)\right) + \lambda \|f\|_{\mcH}^2,\label{eq:kernel_ERM}
\end{equation}
where $\mc L$ is a loss function and $\|f\|_{\mcH}:=\sqrt{\innerH{f,f}}$ is the RKHS norm. By writing $f(x) = \inner{\phi(x), \theta}$ as a parametric function with parameters $\theta\in\mc H$, this optimization over the function space can be written as an optimization over the parameter space as
\begin{align}
    \fkhat(x) &= \inner{\phi(x), \widehat{\theta}}\\
    \widehat{\theta} &= \argmin_{\theta\in \mc H} \sum_{i=1}^n \mc L\left(y_i , \inner{\phi(x_i), \theta}\right) + \lambda \|\theta\|_{L^2}^2.\label{eq:kernel_ERM_params}
\end{align}
Note that this optimization is often very high-dimensional as the dimension of feature space could be very high or even infinite. By the representer theorem \cite{scholkopf2001generalized}, the solution to the optimization problem in \eqref{eq:kernel_ERM} has the form
\begin{equation} \label{eq:fklin}
    \fkhat(x) = \sum_{i=1}^n K(x, x_i)\alpha_i.
\end{equation}
By the reproducing property of the kernel, it is easy to show that $\|\hat f_{\rm ker}\|_{\mcH}^2 = \alpha\tran \Kbf \alpha$ where $\alpha = [\alpha_1, \dots, \alpha_n]\tran$ and $\Kbf$ is the data kernel matrix $\Kbf_{ij} = K(x_i, x_j)$. The optimization problem in \eqref{eq:kernel_ERM} can then be written in terms of $\alpha_i$s as
\begin{equation}
    \widehat{\alpha} = \argmin_{\alpha} \sum_{i=1}^n \mc L(y_i , \Kbf_i \alpha) + \lambda \alpha\tran \Kbf \alpha,\label{eq:kernel_regression_dual}
\end{equation}
where $\Kbf_i$ is the i$th$ row of $\Kbf$.
Observe that this optimization problem only depends on the kernel evaluated over the data points, and hence the optimization problem in \eqref{eq:kernel_ERM} can be solved without ever working in the feature space $\mc H$. If we let $X_\tr$ to represent the data matrix with $x_i$ as its $i$th row, and $y_\tr$ the vector of observations, then for the special case of square loss the optimization problem in \eqref{eq:kernel_regression_dual} has the closed form solution $\widehat{\alpha} = (\Kbf + \lambda I)^{-1} y_\tr$ which corresponds to the estimator
\begin{equation}
    \wh f_{\krr}(x) = K(x, X_\tr) (\Kbf + \lambda I)\inv y_\tr,
\end{equation}
where $K(x, X_\tr)=[K(x,x_1), \dots, K(x,x_n)]$. Throughout this paper, for two matrices of data points $X_1\in \Real^{n_1\times p}$ and $X_2\in\Real^{n_2\times p}$ we use the notation $K(X_1, X_2)$ to represent the $n_1\times n_2$ matrix with $[K(X_1, X_2)]_{ij} = K([X_1]_i, [X_2]_j)$.

\subsection{Gaussian Process Regression}\label{sec:Gaussian_process_regression}
A Gaussian process $f$ is a stochastic process in which for every fixed set of points $\{x_i\}_{i=1}^n$, the joint distribution of $(f(x_1), \dots, f(x_n))$ has multivariate Gaussian distribution. As in multivariate Gaussian distribution, the distribution of a Gaussian process is completely determined by its first and second order statistics, known as the mean function and covariance kernel respectively. If we denote the mean function by $\mu(\cdot)$ and the covariance kernel by $K(\cdot, \cdot)$, then for any finite set of points
\begin{equation}
    \Big( f(x_1), f(x_2),\dots, f(x_n)\Big) \sim \mc N( \mubf, \Kbf),
\end{equation}
where $\mubf$ the vector of mean values $\mubf_i = \mu(x_i)$ and $\Kbf$ is the covariance matrix with $\Kbf_{ij} = K(x_i,x_j)$. Next, assume that a priori we set the mean function to be zero everywhere. Then, the problem of Gaussian process regression can be stated as follows: we are given training samples $\{(x_i, y_i)\}_{i=1}^n$ 
\begin{equation}
    y_i = f(x_i) + \xi_i, \quad \xi_i \stackrel{i.i.d.}{\sim}\mcN(0,\sigma^2),
\end{equation}
where $f$ is a zero mean Gaussian process with covariance kernel $K$. Given a test point $x_\ts$, we are interested in the posterior distribution of $y_\ts:=f(x_\ts)+\xi_\ts$ given the training samples. Defining $X_\tr$ and $y_\tr$ as in previous section we have
\begin{equation}
    \left[\begin{matrix} y_\tr\\y_\ts\end{matrix}\right]|X_\tr, x_\ts\sim \mcN \left(\left[\begin{matrix} 0\\0\end{matrix}\right], \left[\begin{matrix} K(X_\tr, X_\tr) + \sigma^2 I & K(X_\tr, x_\ts)\\K(x_\ts, X_\tr) & K(x_\ts, x_\ts) + \sigma^2\end{matrix}\right]\right),
\end{equation}
where $K(X_\tr, X_\tr)$ is the kernel matrix evaluated at training points. Therefore, if we define $\Kbf:= K(X_\tr, X_\tr)$ we have $y_\ts|y_\tr, X_\tr, x_\ts \sim \mcN(\yhat_\ts, \sigma_\ts^2)$ where
\begin{align}
    \yhat_\ts =& K(x_\ts,X_\tr)(\Kbf + \sigma^2 I)^{-1} y_\tr,\label{eq:conditional_mean}\\
    \sigma_\ts^2 =& \sigma^2 + K(x_\ts, x_\ts) \\
    &- K(x_\ts,X_\tr)(\Kbf + \sigma^2 I)^{-1}K(X_\tr,x_\ts).\label{eq:conditional_variance}
\end{align}
The minimum mean squared error (MMSE) estimator is the estimator that minimizes the square risk
\begin{equation} \label{eq:MMSE_def}
    \widehat{f}_{\rm MMSE} = \argmin_{f\in \mc F}\Exp[(Y_\ts - f(X_\ts))^2|X_\tr, y_\tr],
\end{equation}
where $\mcF$ is the class of all measureable functions of $X$. For a given $x_\ts$, we have $\widehat{f}_{\rm MMSE}(x_\ts) = \widehat{y}_{\rm ts}$ where $\widehat{y}_{\rm ts}$ minimizes the posterior risk
\begin{equation} \label{eq:Emse}
    \Ec(x_\ts)
    := \Exp[\left( \widehat{y}_{\rm ts} - y_{\rm ts} \right)^2|x_\ts, X_\tr, y_\tr]
\end{equation}
and the expectation is with respect to the randomness in $f$ as well as $\{\xi_i\}$. The estimator that minimizes this risk is the mean of the posterior, i.e.\ $\yhat_\ts$ in \eqref{eq:conditional_mean} is the Bayes optimal estimator with respect to mean squared error and its mean squared error is $\Ec(x_\ts) = \sigma_\ts^2$. Note that while this estimator is linear in the training outputs, it is nonlinear in the input data.

In this work, the problem of Gaussian process regression arises for systems that are in the Gaussian kernel regime. More specifically, assume that we have training and test data $\{(x_i, y_i)\}_{i=1}^n$ and $(x_\ts, y_\ts)$ that are generated by a parametric model $y = f(x,\theta) + \xi$ where $\xi \sim \mc N(0,\sigma^2)$. Furthermore, assume that conditioned on $X_\tr$ and $x_\ts$
\begin{equation} \label{eq:fxtrts}
    \left[
    f(x_{\rm ts},\theta), f(x_1,\theta),\ldots,f(x_n,\theta) \right]\tran,
\end{equation}
which is $n+1$-dimensional vector of the function values on the training and test 
inputs is jointly Gaussian and zero mean. Also, for $x$ and $x'$, in the training and test inputs define the kernel function by
\begin{equation}
    K(x,x') := \Exp_{\theta}\left[f(x,\theta)f(x',\theta)\right].
\end{equation}
Then the problem of estimating $\widehat{y}_\ts$ can be considered as a Gaussian regression problem. An important instance of this kernel model is when $f(x,\theta)$ a wide neural network
with parameters $\theta$ drawn from random Gaussian distributions and a linear last layer. In this case, one can show that conditioned on the input, all the preactivation signals in the neural network, i.e.\ all the signals right before going through the nonlinearities, as well as the gradients with respect to the parameters are Gaussian processes as discussed below.

\subsection{Neural Tangent Kernel}\label{app:NTK}
Consider a neural network function $f(x,\theta) = \tilde{\alpha}^{(L)}(x, \theta)$ defined recursively as
\begin{align}
    \alpha^{(0)}(x,\theta) &= x,\\
    \tilde{\alpha}^{(\ell+1)}(x,\theta) &= \frac{1}{\sqrt{n_\ell}}W^{(\ell)}\alpha^{(\ell)}(x,\theta) + \vartheta b^{(\ell)},\\
    \alpha^{(\ell)}(x, \theta) &= \sigma(\tilde{\alpha}^{(\ell)}(x,\theta)), 
\end{align}
where $\sigma$ is a elementwise nonlinearity, $W^{(\ell)}\in \Real^{n_{\ell+1}\times n_{\ell}}$, and $\theta$ is the collection of all weights $W^{(\ell)}$ and biases $b^{(\ell)}$ which are all initialized with i.i.d.\ draws from the standard normal distribution. As noted in many works \cite{neal2012bayesian, lee2017deep, matthews2018gaussian, daniely2016toward}, conditioned on the input signals, with a Lipschitz nonlinearity $\sigma(\cdot)$, the entries of the preactivations $\tilde{\alpha}^{(\ell)}$ converge in distribution to an i.i.d. Gaussian processes in the limit of $n_1, \dots,n_{L\!-\!1}, \red{n_L} \rightarrow \infty$ with covariance $\Sigma^{(\ell)}$ defined recursively as
\begin{align}
    \Sigma^{(1)}(x,x') &= \frac{1}{n_0} x\tran x' + \vartheta^2\\
    \Sigma^{(\ell+1)}(x,x') &= \Exp_{(u,v)\sim \mc N(0, \Sigma^{(\ell)})} \sigma(u)\sigma(v) + \vartheta^2.\label{eq:neural_net_gaussian_kernel}
\end{align}

Therefore, if the true model is a random deep network plus noise, the optimal estimator would be as in \eqref{eq:conditional_mean} with the covariance in \eqref{eq:neural_net_gaussian_kernel} used as the kernel.

The main result of \cite{jacot2018neural} considers the problem of fitting a neural network to a training data using gradient descent. It is shown that in the limit of wide networks (i.e.\ $n_\ell\rightarrow \infty$ for all $\ell$), training a neural network with gradient descent is equivalent to fitting a kernel regression with respect to a specific kernel called the neural tangent kernel (NTK). 

When $f(x,\theta)$ is a neural network with a scalar output, the neural tangent kernel (NTK) is defined as
\begin{equation}
    K(x, x'; \theta) = \inner{\nabla_\theta f(x; \theta), \nabla_\theta f(x'; \theta)}.\label{eq:NTK_def}
\end{equation}
In the limit of wide fully connected neural networks, \cite{jacot2018neural} show that this kernel converges in probability to a kernel that is fixed throughout the training
\begin{equation*}
    K(x, x';\theta) \eqp K(x,x';\theta_0).
\end{equation*}
Similar to \eqref{eq:neural_net_gaussian_kernel}, neural tangent kernel can be evaluated via a set of recursive equations the details of which can be found in \cite{jacot2018neural}. Similar results for architectures other than fully connected networks have since been proven \cite{arora2019exact, yang2019scaling, yang2019wide, alemohammad2020recurrent}.

For a fully connected network with ReLU nonlinearities, the NTK has a closed recursive form given by  \cite{bietti2019inductive}. Let $f(x;\theta) = \sqrt{\frac{2}{n_{L-1}}}\inner{w_{L}, a^{(L-1)}}$ with $a^{(1)} = \sigma(W_1 x)$ and
\begin{equation}
    a^{(\ell)} = \sigma\left(\sqrt{\frac{2}{n_{\ell-1}}}W_\ell a^{(\ell-1)}\right), \quad \ell=2, \dots, L-1,\label{eq:relu_net}
\end{equation}
where $\sigma(\cdot)$ is the ReLU function, $W_\ell \in \Real^{n_\ell \times n_{\ell-1}}$, $w_L\in \Real^{L-1}$ and all the parameters $w_L$ and $W_{\ell}, \ell=1,2, \dots, L-1$, are initialized with i.i.d.\ entries drawn from $\mcN(0,1)$. Then the corresponding NTK, $K(u,v):=K_L(u,v)$ can be obtained recursively by
\begin{align}
    \Sigma_\ell(u,v) &= \|u\|\|v\| \kappa_1\left(\frac{\Sigma_{\ell\!-\!1}(u, v)}{\|u\|\|v\|}\right)\label{eq:NTK_sigma_rec}\\
    K_\ell(u,v)\! &=\! \Sigma_\ell(u,v)\! +\! K_{\ell\!-\!1}(u,v) \kappa_0\left(\frac{\Sigma_{\ell\!-\!1}(u, v)}{\|u\|\|v\|}\right)\label{eq:NTK_K_rec}
\end{align}
for $\ell=1, \dots , L$ and $K_0(u,v)=\Sigma_0(u,v) = u\tran v$ and
\begin{align*}
    \kappa_0(t) &= 1/\pi(\pi - \arccos(t))\\
    \kappa_1(t) &= 1/\pi\left(t\left(\pi - \arccos(t)\right) + \sqrt{1 - t^2}\right).
\end{align*}

%% file: sections/proofs.tex
\section{Generalization of Spectrum of Random Kernel Matrices}\label{sec:proof_of_ElK}
In this section we generalize the results of \cite{el2010spectrum}. In \cite{el2010spectrum}, kernels of the form $K(x_i, x_j)=g(\inner{x_i, x_j}/p)$ are considered whereas here we consider a more general form where $K(x_i, x_j) = g(\tnorm{x_i}^2/p, \inner{x_i, x_j}/p, \tnorm{x_j}^2/p)$. Define $\tau = \lim_{p\rightarrow\infty}\tr{\Sigma_p}/p$. Similar to \cite{el2010spectrum} we assume that 
\begin{itemize}[label=$\bullet$]
    \item $n/p\rightarrow \gamma \in (0,\infty)$ as $p\rightarrow \infty$. 
    \item $x_i = \Sigma_p^{1/2} y_i$ where $y_i\in\Real^p$ has i.i.d.\ entries with $\Exp y_{ik} = 0, \Exp y_{ik}^2 =1$ and $\Exp |y_i|^{4+\varepsilon}<\infty$ for some $\varepsilon>0$.
    \item $\Sigma_p$ is positive definite with bounded operator norm.
    \item $g$ is a $C^3$ function in a neighborhood of $(\tau, 0, \tau)$ and a $C^1$ function in a neighborhood of $(\tau, \tau, \tau)$.
    \item g is symmetric in its first and third argument.
\end{itemize}
\begin{theorem}\label{thm:generalization_of_ELK}
Let $x_i\in \Real^p$ for $i=1, \dots, n$ be $n$ i.i.d. random vectors and form the kernel matrix
\begin{equation*}
    M_{ij} = g\left(\frac{\tnorm{x_i}^2}{p}, \frac{\inner{x_i, x_j}}{p}, \frac{\tnorm{x_j}^2}{p}\right), \quad i,j = 1, \dots, n.
\end{equation*}
Then under the assumptions above we have
\begin{equation*}
    \lim_{p\rightarrow \infty}\tnorm{\Mbf - \Kbf} \eqp 0,
\end{equation*}
where 
\begin{align*}
    \Kbf &= c_0 \Ibf + c_1\ones \ones\tran  +c_3X X\tran %
    c_0 &= g(\tau, \tau, \tau) - g(\tau, 0, \tau) -\frac{\partial f}{\partial z_2}(\tau, 0, \tau)\frac{\tr{\Sigma_p}}{p},\\
    c_1 &= g(\tau, 0, \tau) +\frac{\partial^2 g }{\partial z_2^2}(\tau, 0, \tau)\frac{\tr{\Sigma_p^2}}{2p^2},\\
    c_2 &= \frac{\partial g}{\partial z_2}(\tau, 0, \tau).
\end{align*}
and the partial derivatives are defined for the function $g(z_1, z_2, z_3)$.
\end{theorem}
\begin{proof}
Define $\tau \defeq \lim_{p\rightarrow\infty}\frac{\tr{\Sigma}}{p}$, $z = [\tnorm{x_i}^2/p-\tau, \inner{x_i, x_j}/p, \tnorm{x_j}^2/p]\tran$ and for $i\neq j$ write the second order Taylor expansion of $g(z_1, z_2, z_3)$ around $x_0 = [\tau, 0, \tau]^\top$

\begin{equation*}
    g(\tnorm{x_i}^2/p, \inner{x_i, x_j}/p, \tnorm{x_j}^2/p) = g(x_0) + \inner{\nabla g(x_0), z} + \frac{1}{2} \inner{\nabla^2 g(x_0), z^{\otimes 2}} + R_{ij},
\end{equation*}
where $R_{ij}$ is the Lagrange remainder of this Taylor expansion and has the form
\begin{equation}
    R_{ij} = \frac{1}{6}\sum_{\substack{\alpha_1, \alpha_2, \alpha_3\\ \sum_k\alpha_k =3}}f'''(\xi_1^{ij}, \xi_2^{ij}, \xi_3^{ij}) \left(\frac{\tnorm{x_i}^2}{p}-\tau\right)^{\alpha_1}\left(\frac{\inner{x_i, x_j}}{p}\right)^{\alpha_2}\left(\frac{\tnorm{x_j}^2}{p}-\tau\right)^{\alpha_3},\label{eq:taylor_remainder}
\end{equation}
where $\alpha_k \geq 0$ for some $\xi_1^{ij}, \xi_2^{ij}, \xi_3^{ij}$
\begin{alignat}{2}
    \min(\tau,\tnorm{x_i}^2/p-\tau) &\leq\xi_1^{ij}&&\leq \max(\tau,\tnorm{x_i}^2/p-\tau),\\
    \min(0, \inner{x_i, x_j}/p) &\leq \xi_2^{ij} &&\leq \max(0, \inner{x_i, x_j}/p),\\
    \min(\tau,\tnorm{x_j}^2/p-\tau) &\leq\xi_3^{ij}&&\leq     \max(\tau,\tnorm{x_j}^2/p-\tau).
\end{alignat}
As is shown in \cite{el2010spectrum}, under the Assumptions A1-A3, for some $0<\delta<1/2$ we have 
\begin{equation}
    \max_i \left|\|x_i\|_2^2/p - \tau\right|\leq p^{-\delta/2}, \quad \max_{i\neq j} |\inner{x_i, x_j}/p| \leq p^{-\delta}\log p. 
\end{equation}
Therefore, we have
\begin{equation}
    \max_{i\neq j} |\xi_k^{ij} -\tau| \rightarrow 0 \quad\text{a.s. for $k=1, 3$},
\end{equation}
and
\begin{equation}
    \max_{i\neq j} |\xi_2^{ij}|\rightarrow 0 \quad \text{a.s.}
\end{equation}
Therefore, by continuity assumptions $f'''(\xi_1^{ij}, \xi_2^{ij}, \xi_3^{ij})\rightarrow f'''(\tau, 0, \tau)$ and it is also bounded. Therefore, a similar argument to \cite{el2010spectrum}, shows that ignoring the terms that have $(\tnorm{x_i}^2/p - \tau)$ would result in a consistent approximation of the kernel matrix in operator norm. In particular, the matrices corresponding to all the terms in the Taylor expansion remainder in \eqref{eq:taylor_remainder} except for the matrix with entries
\begin{equation}
    \tilde{R}_{ij} = \left(\frac{\inner{x_i, x_j}}{p}\right)^3
\end{equation} 
for $i\neq j$ and zeros on the diagonal can be ignored. But this matrix has exactly the form that is considered in \cite{el2010spectrum} and it is shown that this matrix too can be ignored in operator norm in the limit. Therefore, the error of the Taylor expansion can be ignored so long as consistency in operator norm is desired.

Ignoring all the terms that involve $(\tnorm{x_i}^2/p - \tau)$ in the Taylor expansion for $i\neq j$ we get
\begin{equation*}
    g(\tnorm{x_i}^2/p, \inner{x_i, x_j}/p, \tnorm{x_j}^2/p) = g(x_0) + \frac{\partial g}{\partial z_2}\frac{\inner{x_i, x_j}}{p} + \frac{1}{2} \frac{\partial^2 g}{\partial z_2^2}\left(\frac{\inner{x_i, x_j}}{p}\right)^2,
\end{equation*}
results in a consistent approximation of the kernel matrix in operator norm.

For $i=j$, we have that
\begin{equation}
    K(x_i ,x_i) = g\left(\frac{\tnorm{x_i}^2}{p}, \frac{\tnorm{x_i}^2}{p},\frac{\tnorm{x_i}^2}{p} \right),
\end{equation}
which is exactly of the form considered in \cite{el2010spectrum}. Therefore, we can use the same approximation for the diagonal entries of the kernel matrix. Putting everything together, the result is proved.

\end{proof}

\subsection{Proof of Theorem \ref{thm:equivalence_linear_kernel}}\label{sec:proof_of_equivalence}
\begin{proof}
Let $\overline{X}=[x_\ts\tran, X_\tr\tran]\tran$ and partition the kernel matrix, 
$K(\overline{X},\overline{X})$ as 
\begin{align}
    K(\overline{X},\overline{X}) &= \left[ \begin{array}{cc}
         K(x_{\rm ts},x_{\rm ts}) & K(x_{\rm ts},X_{\rm tr})  \\
         K(X_{\rm tr},x_{\rm ts}) & K(X_{\rm tr},X_{\rm tr}) 
    \end{array}\right]\\ &=: 
    \left[ \begin{array}{cc}
         M_{11}  & M_{12}  \\
         M_{21} & M_{22} 
    \end{array}\right] =: M,
\end{align}
where we have used $M_{ij}$, $i,j=1,2$ to denote the four components
of $K(\overline{X},\overline{X})$.  With this notation,
the optimal estimator in \eqref{eq:kernel_estimator} is
\begin{equation} \label{eq:yhatlsm}
    \fkhat(x_\ts)= M_{12}(M_{22} + \lambda I)^{-1}y_\tr.
\end{equation}
Next, use the approximation of $K(\overline{X}, \overline{X})$ from Proposition \ref{thm:generalization_of_ELK_result}
\begin{equation} \label{eq:Mhat}
    \widehat{K} := c_0 I  + c_1 11\tran + \frac{c_2}{p} \overline{X}
    \overline{X}\tran,
\end{equation}
where $I$ is the $(n+1)\times (n+1)$ identity matrix and $1$ is an all one vector of size $n+1$.
The result of Proposition \ref{thm:generalization_of_ELK_result} states that
\begin{equation} \label{eq:Mhatlim}
    \lim_{n \rightarrow \infty} \| K - \widehat{K} \| \eqp 0.
\end{equation}
Now, partition the matrix $\widehat{K}$ in \eqref{eq:Mhat} as,
\begin{equation} \label{eq:Mhatpart}
    \widehat{K} = \left[ \begin{array}{cc}
         \widehat{K}_{11}  & \widehat{K}_{12}  \\
         \widehat{K}_{21} & \widehat{K}_{22} 
    \end{array}\right].
\end{equation}
We first show that each of the blocks in this matrix converge to the corresponding block in $K$ in operator norm (or $\ell_2$ norm for when the block is a vector). By another application of Proposition \ref{thm:generalization_of_ELK_result}, we have that
$\lim_{n\rightarrow\infty}\tnorm{K_{22} - \widehat{K}_{22}}\eqp 0$. Next, consider the vector $e_1 = [1, 0, 0, ..., 0]\tran \in\Real^{n+1}$. We have
\begin{align}
    \left\|\left[ \begin{array}{cc}
         K_{11} - \widehat{K}_{11} \\
        K_{21} -  \widehat{K}_{21}
    \end{array}\right]\right\|_2 &=\left\|(K-\widehat{K})e_1\right\|_2\\
    &\leq \tnorm{K-\widehat{K}}\tnorm{e_1}\\
    & \eqp 0.
\end{align}
Therefore, $\lim_{n\rightarrow \infty}|K_{11} - \widehat{K}_{11}|\eqp 0$ and $\lim_{n\rightarrow \infty}\|K_{21} - \widehat{K}_{21}\|_2\eqp 0$.

Next, consider the estimator in \eqref{eq:yhatlsm} with $M$ replace by $\widehat{K}$
\begin{equation}
    \tilde{f}_{\rm ker}(x_\ts) =  \widehat{K}_{12}(\widehat{K}_{22} + \lambda I)^{-1} y_{\rm tr}.
\end{equation}
By \eqref{eq:Mhatlim}, we have that
\begin{equation}
    \lim_{n\rightarrow \infty} |\tilde{f}_{\rm ker}(x_\ts) - \fkhat(x_\ts)| \eqp 0.\label{eq:equivalence_of_kernel_and_kernel_approx}
\end{equation}
Next we show that with the selection of the regularization parameters in \eqref{eq:optimal_reg}, the linear estimator in \eqref{eq:yhatls_solution} is also equal to $\tilde{f}_{\rm ker}(x_\ts)$, i.e. the linear estimator in Model (2) is the same as a kernel estimator with the approximation in \eqref{eq:Mhat} used as the kernel.

Using \eqref{eq:Mhat} we have
\begin{align*}
    \widehat{K}_{11} &= c_0 + c_1 + x_\ts x_\ts\tran,\\
    \widehat{K}_{12} &= c_1 1\tran + \frac{c_2}{p}x_\ts X_\tr\tran,\\
    \widehat{K}_{22} &= c_0 I + c_1 1\tran 1 + \frac{c_2}{p}X\tran_\tr X_\tr.
\end{align*}
Therefore,
\[
    \tilde{f}_{\rm ker}(x_\ts) = \left[c_1 1\tran + \frac{c_2}{p}x_\ts X_\tr\tran\right]
    \left[c_1 1 1\tran + \frac{c_2}{p}X_\tr X_\tr\tran + (c_0+\lambda) I\right]^{-1}y_\tr.\label{eq:yhat_Mhat}
\]
Comparing \eqref{eq:yhat_Mhat} with \eqref{eq:yhatls_solution} we see that each of the terms in the linear model $f_{\rm lin}(x_\ts)$ and the corresponding term in kernel model $\tilde{f}_{\rm ker}(x_\ts)$ are equal in probability when
\begin{equation}
    \lambda_1 = \frac{c_0 + \lambda}{c_1}, \quad \lambda_2 = \frac{p(c_0 + \lambda)}{c_2}.
\end{equation}
Therefore, using Lemma \ref{lem:convergence_of_product} we see that their product should also converge in probability to each other.
In other words, with this choice of regularization parameters
\begin{equation}
    \lim_{n\rightarrow \infty} |\widehat{f}_{\rm lin}(x_\ts) - \tilde{f}_{\rm ker}(x_\ts)| \eqp 0.\label{eq:equivalence_of_linear_and_kernel_approx}
\end{equation}
Putting \eqref{eq:equivalence_of_kernel_and_kernel_approx} and \eqref{eq:equivalence_of_linear_and_kernel_approx} we get the desired results
\begin{equation}
    \lim_{n\rightarrow \infty} |\widehat{f}_{\rm lin}(x_\ts) - \fkhat(x_\ts)| \eqp 0.
\end{equation}
\end{proof}

\section{Technical Lemmas}\label{app:technical}
\begin{lemma}\label{lem:matrix_inversion} Let A be an invertible $n\times n$ matrix, and $U\in\Real^{d\times n}$ for some d, then 
\begin{align}
    (A+UU\tran)\inv U = A\inv U(I_n + U\tran A\inv U)\inv C\inv
\end{align}

\end{lemma}

\begin{lemma}\label{lem:convergence_of_product}
Let $A_i\in\Real^{n_1\times n_2}$ and $B_i\in\Real^{n_2\times n_3}$ be two sequences of random matrices and assume that 
\begin{equation}
    \lim_{i\rightarrow\infty} \tnorm{A_i-A} \eqp  0, \quad \lim_{i\rightarrow\infty} \tnorm{B_i-B} \eqp 0.
\end{equation}
Then if $\tnorm{A},\tnorm{B}<\infty$ we have
\begin{equation}
    \lim_{i\rightarrow\infty} \tnorm{A_iB_i - AB}\eqp 0.
\end{equation}
\end{lemma}
\begin{proof}
\begin{align}
     \tnorm{A_iB_i - AB} &= \tnorm{A_i B_i - A B_i + AB_i -AB}\\
    &\leq \tnorm{A_i B_i - A B_i} + \tnorm{ AB_i -AB}\\
    & \leq \tnorm{A_i - A}\tnorm{B_i} + \tnorm{A}\tnorm{B_i - B}\\
    &\eqp 0,
\end{align}
where the last equality follows from the continuous mapping theorem (\cite{mann1943stochastic}). This proves the claim.
   
\end{proof}
A special case of this theorem is when $B_i$ is a sequence of $n_2\times 1$ matrices, i.e.\ a sequence of vectors. In this case, the operator norm is the same as the $\ell_2$ norm. Therefore, we have the following corollary.
\begin{corollary}
Let $A_i\in\Real^{n_1\times n_2}$ and $x_i\in\Real^{n_2\times n_3}$ be a sequence of random matrices and random vectors respectively and let
\begin{equation}
    \lim_{i\rightarrow\infty} \tnorm{A_i-A} \eqp  0, \quad \lim_{i\rightarrow\infty} \tnorm{x_i-x} \eqp 0.
\end{equation}
Then if $\tnorm{A},\tnorm{x}<\infty$ we have
\begin{equation}
    \lim_{i\rightarrow\infty} \tnorm{A_ix_i - Ax}\eqp 0.
\end{equation}
\end{corollary}
The next corollary considers limits of powers of a matrix which can be proven by a simple induction using Lemma \ref{lem:convergence_of_product}.
\begin{corollary}
Let $A_i\in\Real^{n\times n}$ be a sequence of random matrices and assume that $\lim_{i\rightarrow \infty}\tnorm{A_i - A}\eqp 0$. Then for any finite $m\in \mathbb{N}$ we have
$\lim_{i\rightarrow \infty} \tnorm{A_i^m - A^m}\eqp 0$.
\end{corollary}
\section{Proof of Theorem \ref{thm:equivalence_GD}}\label{sec:proof_of_equivalence_GD}
Here we show that if the kernel model and linear model are learned by gradient descent, they are equivalent to each other throughout the training.

Consider a kernel model parameterized in the feature space
\begin{align}
    \fkhat(x) &= \inner{\phi(x), \widehat{\theta}}\\
    \widehat{\theta} &= \argmin_{\theta} \left\|(y_\tr - \phi(X_\tr)\theta\right\|_2^2 + \lambda \|\theta\|_{L^2}^2,
\end{align}
where $\phi(X_\tr)$ is a matrix with $\phi(x_i)\tran$ as its $i$th row.
The gradient descent update for this problem is
\begin{equation}
    \theta^{t+1} = (I -\eta ((\phi(X_\tr)\tran \phi(X_\tr) + \lambda I))\theta^t + \eta \phi(X_\tr)\tran y_\tr,
\end{equation}
where $\eta$ is the learning rate. Therefore, if initialized with $\theta_0=0$, after $t$ steps of gradient descent we obtain
\begin{equation}
    \theta^t = \sum_{t'=0}^{t-1}\eta(I -\eta ((\phi(X_\tr)\tran \phi(X_\tr) + \lambda I))^{t'}\phi(X_\tr)\tran y_\tr.\label{eq:gd_kernel}
\end{equation}
In order to further simplify this result, we use the following lemma.
\begin{lemma}\label{lem:push_through}
For any integer $t' \geq 0$ and matrix $A\in\Real^{n\times p}$ we have
\begin{equation}
    (\alpha I_{p\times p} - A\tran A)A\tran = A\tran (\alpha I_{n\times n} - A A\tran).
\end{equation}
\begin{proof}
This result can be easily proved by using the singular value decomposition of $X=U\Sigma V\tran$.
\end{proof}
\end{lemma}
Using this lemma we have
\begin{equation}
    \theta^t =\sum_{t'=0}^{t-1}\eta\phi(X_\tr)\tran(I -\eta ((\phi(X_\tr) \phi(X_\tr)\tran + \lambda I))^{t'}y_\tr.
\end{equation}
Note that the identity matrix in this equation has a different size from the one in \eqref{eq:gd_kernel} and the sizes can be inferred from the number of samples as well as dimension of the feature space as in the Lemma \ref{lem:push_through}. Therefore, by observing that $K(x_i,x_j) = \inner{\phi(x_i), \phi(x_j)}$ the model at time $t$ represented by $\fkhat^t$ evaluated on test data point $x_\ts$ has the form
\begin{align}
    \fkhat^t(x_\ts) &= \eta K(x_\ts, X_\tr)\sum_{t'=0}^{t-1}(I -\eta ((K(X_\tr, X_\tr) + \lambda I))^{t'}y_\tr\\
    & = K(x_\ts, X_\tr)((K(X_\tr, X_\tr) + \lambda I)^{-1}\bigg(I - 
    \big(I -\eta ((K(X_\tr, X_\tr) + \lambda I)\big)^{t}\bigg)y_\tr\label{eq:kernel_gd_solution}
\end{align}
This series and hence the gradient descent converges if all the eigenvalues of $I -\eta ((K(X_\tr, X_\tr) + \lambda I)$ lie inside the unit circle which is always possible by choosing $\eta$ that is small enough and the limiting solution is the kernel regression solution
\begin{equation}
    \widehat{y}_\ts = K(x_\ts, X_\tr)((K(X_\tr, X_\tr) + \lambda I)^{-1}y_\tr.
\end{equation}
Now consider a scaled linear model $f_{\rm lin}(x) = \gamma_2 w\tran x + \gamma_1 b$ fitted applying gradient descent to $\ell_2$-regularized least squares loss, i.e.
\begin{align}
    (\widehat{w},\widehat{b}) =& \argmin_{w,b} J(w,b),  \\
    J(w,b) :=& \sum_{i=1}^n (y_i - \gamma_2 w\tran x_i - \gamma_1 b)^2  + \lambda_1 |\gamma_1 b|^2 + \lambda_2 \|\gamma_2w\|^2.
\end{align}
The gradient descent update for this problem is
\begin{equation}
    \left[\begin{matrix}w^{t+1}\\b^{t+1}\end{matrix}\right] =
    \left(I_{(p+1)\times (p+1)} - \eta \left[\begin{matrix}\gamma_2^2 X_\tr\tran X_\tr +\gamma_2^2\lambda_2 I_{p\times p} & \gamma_2\gamma_1 X_\tr\tran 1\\\gamma_2\gamma_1 1\tran X_\tr & \gamma_1^2n +\gamma_1^2\lambda_1\end{matrix}\right]\right)\left[\begin{matrix}w^{t}\\b^{t}\end{matrix}\right] + \eta \left[\begin{matrix}\gamma_2X_\tr\tran\\\gamma_1 1\tran\end{matrix}\right]y_\tr,
\end{equation}
where $\eta$ is the learning rate. Next, set $\gamma_1 = \gamma_2\sqrt{\lambda_2/\lambda_1}$ such that $\gamma_1^2\lambda_1 = \gamma_2^2\lambda_2$. Then, if $w^0 = 0$ and $b^0 = 0$, at time $t$ we have
\begin{equation}
    \left[\begin{matrix}w^{t}\\b^{t}\end{matrix}\right] =
    \sum_{t'=0}^{t-1}\eta\left(I_{(p+1)\times (p+1)} - \eta \left[\begin{matrix}\gamma_2^2X_\tr\tran X_\tr +\gamma_2^2\lambda_2 I_{p\times p} & \gamma_1\gamma_2 X_\tr\tran 1\\\gamma_1 \gamma_2 1\tran X_\tr & \gamma_1^2n +\gamma_2^2\lambda_2\end{matrix}\right]\right)^{t'}\left[\begin{matrix}\gamma_2X_\tr\tran\\\gamma_1 1\tran\end{matrix}\right]y_\tr,
\end{equation}
Now we can apply Lemma \ref{lem:push_through} with
\begin{equation}
    A = [\gamma_2X_\tr, \gamma_1 1], \quad \alpha =\lambda_2,
\end{equation}
to get
\begin{equation}
    \left[\begin{matrix}w^{t}\\b^{t}\end{matrix}\right] =
    \sum_{t'=0}^{t-1}\eta\left[\begin{matrix}\gamma_2X_\tr\tran\\\gamma_1 1\tran\end{matrix}\right]\left(I_{n\times n} - \eta (\gamma_2^2X_\tr X_\tr \tran + \gamma_1^2 1 1\tran +\gamma_2^2\lambda_2 I_{n\times n})\right)^{t'}y_\tr.
\end{equation}
Therefore, the linear model $f_{\rm lin}(x) = \gamma_2w\tran x + \gamma_1 b$  when evaluated on a test data $x_\ts$ after $t$ step of gradient descent $f^t_{\rm lin}$ takes the form
\begin{align}
    f^t_{\rm lin}(x_\ts) &=\sum_{t'=0}^{t-1}\eta(\gamma_2^2x_\ts X_\tr \tran + \gamma_1^2 1\tran)\left(I_{n\times n} - \eta (\gamma_2^2X_\tr X_\tr \tran + \gamma_1^2 1 1\tran +\gamma_2^2 \lambda_2 I_{n\times n})\right)^{t'}y_\tr\\
    & = (\gamma_2^2x_\ts X_\tr \tran + \gamma_1^2 1\tran)(\gamma_2^2X_\tr X_\tr \tran + \gamma_1^2 1 1\tran +\gamma_2^2\lambda_2 I_{n\times n})\inv\\&\hspace{8em}\left(I_{n\times n} - \left(I_{n\times n} - \eta (\gamma_2^2X_\tr X_\tr \tran + \gamma_1^2 1 1\tran +\gamma_2^2\lambda_2 I_{n\times n})\right)^{t}\right)y_\tr.
\end{align}
Now, recall the approximation of data kernel matrix in \eqref{eq:Mhat}
\begin{equation}
    \left\|\left[ \begin{array}{cc}
         K(x_{\rm ts},x_{\rm ts}) & K(x_{\rm ts},X_{\rm tr})  \\
         K(X_{\rm tr},x_{\rm ts}) & K(X_{\rm tr},X_{\rm tr}) 
    \end{array}\right] -  \left[ \begin{array}{cc}
         \widehat{K}_{11}  & \widehat{K}_{12}  \\
         \widehat{K}_{21} & \widehat{K}_{22} 
    \end{array}\right]\right\|_{\rm op}\eqp 0,
\end{equation}
where 
\begin{align*}
    \widehat{K}_{11} &= c_0 + c_1 + x_\ts x_\ts\tran,\\
    \widehat{K}_{12} &= c_1 1\tran + \frac{c_2}{p}x_\ts X_\tr\tran,\\
    \widehat{K}_{22} &= c_0 I + c_1 1\tran 1 + \frac{c_2}{p}X\tran_\tr X_\tr.
\end{align*}
Then, using $\gamma_1^2 = \gamma_2^2 \lambda_2/\lambda_1$ and $\gamma_2^2 = c_2/p$ and the regularization parameter $\lambda_1 = (c_0+\lambda)/c_1$ and $\lambda_2 = p(c_0+\lambda)/c_2$ we get
\begin{multline}
    f^t_{\rm lin}(x_\ts) = (\frac{c_2}{p}x_\ts X_\tr \tran + c_1 1\tran)(\frac{c_2}{p}X_\tr X_\tr \tran + c_1 1 1\tran +(c_0 + \lambda) I_{n\times n})\inv\\\left(I_{n\times n} - \left(I_{n\times n} - \eta (\frac{c_2}{p} X_\tr X_\tr \tran + c_1 1 1\tran +(c_0 + \lambda) I_{n\times n})\right)^{t}\right)y_\tr.
\end{multline}
As in the proof of Theorem \ref{thm:equivalence_linear_kernel}, Each of these terms converges in probability to the corresponding term in \eqref{eq:kernel_gd_solution}. Therefore, applying Lemma \ref{lem:convergence_of_product}, for any $t \geq 0$ we have
\begin{equation}
    {\fkhat}^t(x_\ts) \eqp f_{\rm lin}^t(x_\ts).
\end{equation}